# Evolutionary Approaches to Expensive Optimisation

Maumita Bhattacharya
School of Computing & Mathematics
Charles Sturt University
NSW, Australia - 2640

*Abstract*— Surrogate assisted evolutionary algorithms (EA) are rapidly gaining popularity where applications of EA in complex real world problem domains are concerned. Although EAs are powerful global optimizers, finding optimal solution to complex high dimensional, multimodal problems often require very expensive fitness function evaluations. Needless to say, this could brand any population-based iterative optimization technique to be the most crippling choice to handle such problems. Use of approximate model or surrogates provides a much cheaper option. However, naturally this cheaper option comes with its own price! This paper discusses some of the key issues involved with use of approximation in evolutionary algorithm, possible best practices and solutions. Answers to the following questions have been sought: what type of fitness approximation to be used; which approximation model to use; how to integrate the approximation model in EA; how much approximation to use; and how to ensure reliable approximation.

*Keywords—Optimization; Evolutionary Algorithm, Approximation Model; Fitness Approximation; Meta-model; Surrogate*

I. INTRODUCTION

Evolutionary algorithms (EAs) have long been accepted as powerful search algorithms, with numerous applications in various science and engineering problem domains. Evolutionary algorithms deserve a special mention as powerful global optimizers. Also, evolutionary algorithms are found to outperform conventional optimization algorithms in problem domains involving, discontinuous, non-differential, multi-modal, noisy, and not well-defined problems. However, many real world optimization problems including engineering design optimization often involve computationally very expensive function evaluations. This makes it impractical for a population-based iterative search technique such as Evolutionary Algorithm (EA) to be used in such problem domains. The runtime for a single function evaluation, in such problems, could be in the range from a fraction of a second to hours of supercomputer time. A viable alternative is to use approximation instead of actual function evaluation to substantially reduce the computation time [39, 50 and 51].

Use of surrogates to speed up optimization is not a new concept [6-15]. The earliest trials date back to the sixties. The most widely used models being Response Surface Methodology [47], Krieging models [55] and artificial neural network models [16]. In the multidisciplinary optimisation (MDO) community, primarily response surface analysis and polynomial fitting techniques are used to build the approximate models [26, 59]. These models work well when single point traditional gradient-based optimisation methods are used.

However, they are not well suited for high dimensional multimodal problems as they generally carry out approximation using simple quadratic models. In another approach, multilevel search strategies are developed using special relationship between the approximate and the actual model. An interesting class of such models focuses on having many islands using low accuracy/cheap evaluation models with small number of finite elements that progressively propagate individuals to fewer islands using more accurate but expensive evaluations [60]. As is observed in [32], this approach may suffer from lower complexity, cheap islands having false optima whose fitness values are higher than those in the higher complexity, expensive islands. Rasheed et al. in [50, 51], uses a method of maintaining a large sample of points divided into clusters. Least square quadratic approximations are periodically formed of the entire sample as well as the big clusters. Problem of unevaluable points was taken into account as a design aspect. However, it is only logical to accept that true evaluation should be used along with approximation for reliable results in most practical situations. Another approach using population clustering is that of fitness imitation [32]. Here, the population is clustered into several groups and true evaluation is done only for the cluster representative [39]. The fitness value of other members of the same cluster is estimated by a distance measure. The method may be too simplistic to be reliable, where the population landscape is a complex, multimodal one.

Jin et al. in [36 and 34] analysed the convergence property of approximate fitness-based evolutionary algorithm. It has been observed that incorrect convergence can occur due to false optima introduced by approximate models. Two controlled evolution strategies have been introduced. In this approach, new solutions (offspring) can be (pre)-evaluated by the model. The (pre)-evaluation can be used to indicate promising solutions. It is not clear however, how to decide on the optimal fraction of the new individuals for which true evaluation should be done [17]. In an alternative approach, the optimum is first searched on the model. The obtained optimum is then evaluated on the objective function and added to the training data of the model [52, 58, and 17]. Yet in another approach as proposed in [36], a regularization technique is used to eliminate false minima.

Although using regression and interpolation tools such as least square regression, back propagating artificial neural network, response surface models, and so on are effective means for building the approximate models, accuracy of the result is a major risk involved in using meta-models to replace actual function evaluation [32, 36, 34 and 59]. Fig. 3 depicts how levels of fitness evaluations influence computational expense and accuracy..

Apart from the type of the meta-model generator used, the concepts of using approximate model vary (i) in approximation strategies i.e., what exactly is approximated, (ii) in the model integration mechanism used, and (iii) in model management techniques used [32]. This paper discusses some of these crucial aspects of surrogate assisted evolutionary algorithms.

The rest of the paper is organised as follows. Section II briefly outlines the key issues involved with surrogate assisted evolutionary algorithms; Section III presents the different approximation strategies or types of approximation; Section IV briefly mentions the commonly used approximation model generation tools; while Section V discusses the approximation model integration mechanisms. Section VI and Section VII respectively discusses how much approximation to be used and the issue of quality assurance while using meta-models. Section VIII presents some concluding remarks.

## II. Issues Involved With Surrogate Assisted EA

Replacing actual analysis or evaluation by approximate model involves risks and several issues need to be addressed in employing fitness approximations in evolutionary computation (Fig. 1). Of the several issues, foremost are:

- What type of fitness approximation to be used;
- Which approximation model to use;
- How to integrate the approximation model in EA;
- How much approximation to use;
- How to ensure reliable approximation.

Fitness evaluation can be performed by experimental evaluation, complete computational simulation, simplified computational simulation as well as by approximation with surrogates or meta-models; while experimental evaluation can be treated as the true fitness value of a given candidate solution. The tradeoff between computational expense and accuracy is as depicted in Fig. 3. Quite naturally, actual experimental evaluation of fitness gives the highest accuracy but incurs the highest computational cost as well. Fitness evaluation by approximation with surrogates is order of magnitude cheaper compared to the other techniques; but, it also results in lowest accuracy.

Due to inadequate amount of data, ill sampling and the high dimensionality of data sets (input space), it is often very difficult to obtain an *accurate* global approximation of the original fitness function. Hence, the approximate model should be used together with the true fitness function. In most cases, the original fitness function is available, although it is computationally very expensive. Therefore, it is only feasible to use the original fitness function sparingly. The mechanism controlling *how much* of expensive evaluation should be incorporated and in *what way*, is known as *model management* in conventional optimization [21] or *evolution control* in evolutionary computation literature [41, 34]. Also, considering the limited number of sample points that can be available, the quality of the approximate model could be improved by intelligent model selection, use of active data sampling and on-line and off-line weighting, selection of training method and selection of error measures.

Some of these issues related to using approximate model or surrogate in evolutionary algorithm are detailed in the following sections.

## III. Types of Approximation

There are various strategies to use approximation in optimization problems. Two such more traditional approaches are [32]: problem approximation and functional approximation. A number of other specialized approaches have been implemented for evolutionary fitness evaluation.

### A. Problem Approximation

In this approach, the statement of the problem itself is replaced by a reduced one that is easier to solve. One such example is reported in [5], where, in CFD simulations, the fluid dynamics are described with three-dimensional (3D) Navier-Stokes equations with a turbulence model. Subjected to certain constraints, the 3D flow field can be solved by 2D computations, which is computationally less expensive. Some other examples are reported in [24, 3].

### B. Functional Approximation

As the name suggests, in this approach, an alternate and explicit expression is constructed for the objective function, for the purpose of reducing the cost of evaluation.

The surrogate assisted EA techniques reported in [6,8, 10, 12 and 14] uses approximate models to evaluate fitness to reduce the number of actual fitness evaluation. Refer to [32] for more examples on the functional approximation technique.

### C. EA Specififc Approximation

This approach is specific for evolutionary algorithms and utilizes the algorithm's structural and functional aspects. *Fitness inheritance* is an example of this technique. In this approach, fitness value of the offspring is estimated from the fitness value of the parents to reduce actual fitness evaluations. In an alternative approach called *fitness imitation*, the individuals are clustered into several groups. Then, only the representative individual of the clusters are evaluated using expensive fitness evaluation. The fitness values of the remaining individuals in the cluster are estimated based on the actual fitness value of the representative individual. Fitness inheritance/ fitness imitation has been used in several researches [66, 56, 18 and 44].

Which of the above three types of approximation should be used in a specific case, naturally depends on the actual intent of using surrogates in the first place.

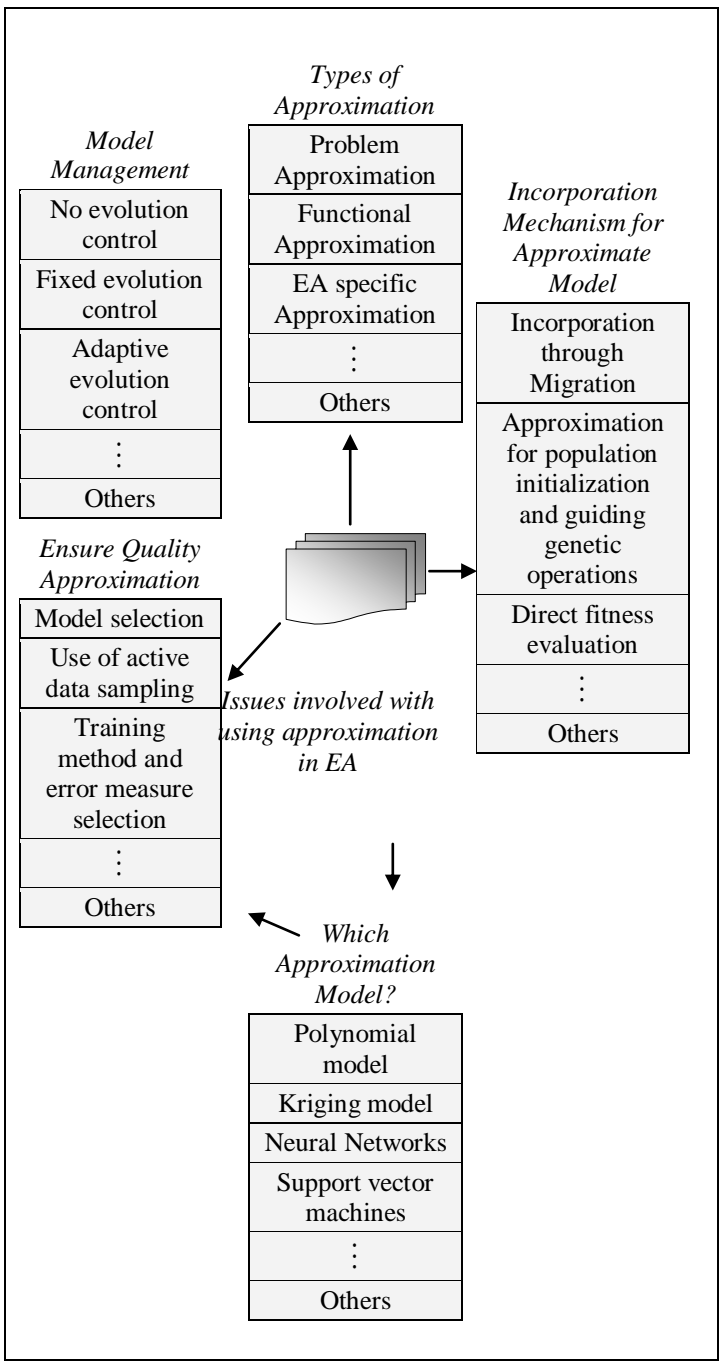

Fig. 1. The issues involved with using approximation in evolutionary algorithm.

IV. APPROXIMATION MODEL GENERATION TOOLS

Approximate models or meta-models in this context are models that are developed to approximate computationally expensive simulation codes. Functional approximation modeling generally involves finding a set of parameters for a given model to find the good, best or perfect fit between a given finite sampling of values of independent variables and associated values of dependent variables [32]. A wide variety of empirical tools are used to generate functional approximation models. Some of the commonly used ones are polynomial interpolation, DACE (design and analysis of computer model) or kriging model, artificial neural networks, regression spline etc. An important characteristic of a meta model generator is generalization. Generalization is the ability to map or predict values that were not considered in the training set while developing the model. The least square method (LSM) performs efficiently only within a small trust region and fails in terms of generalization particularly for complex polynomials with discontinuity in the target function. However, for low dimensional problems with real valued parameters, the polynomial regression models often outperform the connectionist methods. The connectionist models, like the neural networks perform better for high dimensional problems. Unlike the LSM, the kriging models are capable of capturing multiple local extrema, but at the expense of higher computational cost.

It is hard to compare the performances of the different model approximators as performance can be problem dependent and also there are several criteria that need to be considered. However, the most important ones are the accuracy, both on the training and the test data, computational complexity and transparency [32]. One of the serious problems is the introduction of false optima. A desirable tradeoff may be that of lower approximation accuracy if the model is used in global optimization. Some methods for prevention of false minima in neural network are available.

In [32] Jin has suggested the following general rules for model selection. It is recommended to implement first a simple approximate model, for example, a lower order polynomial model to see if the given samples can be fitted with reasonable accuracy. If it fails a model with higher complexity such as higher order polynomials or neural network models should be considered, However, for high dimensional problems with small number of samples, a neural network model is generally preferable. In case of neural network models, in particular a multilayer perceptrons network, the model complexity should be controlled to avoid over-fitting. The gradient descent based method might lead to slow convergence in some cases. The RBF networks show superior performance both in terms of accuracy and training speed for some problems. Support vector machine based approximators, on the other hand, are known to provide robust performance in high dimensional problems with fewer samples.

For further information on non EA specific surrogate assisted design and analysis, see [1, 29].

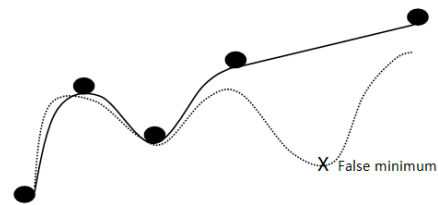

Fig. 2. An example of a false minimum in the approximate model. The solid curve denotes the original function and the dashed curve denotes its approximation.

## V. Approximation Model Integration Mechanisms

Integration of approximate model in EA can be done at various levels; e.g., population initialization, with genetic operators such as recombination and mutation, migration in multi-population architecture and so on. The major ones are as discussed below.

- Fitness approximation has been used to initialize population and to guide genetic operators (recombination and mutation) in [4, 50 and 2]. Using approximation only for initialization and for guiding genetic operators is expected to reduce the associated risk of using approximate model as such operations are required only occasionally. However, reduction in actual fitness evaluation may not be that significant [32].

- Yet another approach directly uses approximation based fitness evaluation in order to reduce the number of actual fitness evaluations. This approach of incorporation of approximation is of interest to us in connection with the frameworks proposed in this paper. Different approaches have been proposed with varied degrees of success. Some of the works are reported in [42, 46, 52, 34; 36]. Application to multiobjective optimization has been reported in [23, 48 and 49]. Different approximation model generators and approximation control to some degree have been proposed.

- Approximation with migration may be implemented by maintaining sub-populations at different levels of approximation and allowing migration of individuals from one level of sub-population to another based on pre-defined rules. This has been implemented in [60, 57 and 22].

## VI. How Much Approximation?

In the context of reducing the number of actual fitness evaluations, among the various approaches to incorporate approximate models (see Section IV for description of the integration mechanisms), using approximate models for fitness evaluations is most effective. In the real world it is quite common not to have any clear analytical fitness function to accurately compute the fitness of a candidate solution. Depending on the level of estimation used, the compromise between accuracy and computational cost is achieved (see Fig. 3).

Nonetheless, any mechanism to use approximation in EA should try to achieve the following:

- The evolutionary algorithm should converge to the global optimum or at least to a near optimum of the original function. However, in reality it is very difficult to construct such an approximate model due to high dimensionality of the problem, inadequate number of training samples and poor distribution of the candidate solutions in the search space. It is obvious that with some form of approximation control, it is very likely that the evolutionary algorithm will converge to a false optimum introduced by the approximate model. See Fig.2 for an example.

- The overhead of maintaining the approximation model/models should be kept low so that the expenses do not outweigh the benefits.

Using true fitness evaluation along with approximation is thus extremely important to achieve reliable performance by the surrogate assisted EA mechanism. This can be regarded as the issue of model management or evolution control [41, 34].

In the simplest form of model management true function evaluation is not used at all [37, 53]. This is feasible only if the approximate model is considered to be of high fidelity. In most cases, however, evolution control or model management must be used. Some of the popular ones are as follows.

- Surrogates may be used in some of the generations only instead of in all generations of the evolutionary process. Some of the examples are [40, 36 and 20].

- In another approach, surrogates may be used for specific individuals in a generation/ generations only instead of for the entire population. See [27, 36].

- In yet another approach, more than one sub-population may co-evolve using their own surrogate model for fitness evaluation. Migration from one such population to another can occur.

- Specialized model management methods may be necessary for some surrogated assisted evolutionary algorithms [64, 65 and 54]. [65] uses the method for single objective optimization and [54] for multi-objective optimization. Adjusting the frequency of evolution control according to the reliability of the approximate model seems logical [33]. Along with a generation-based approach, [48] has suggested a method to adjust the frequency of evolution control based on the trust region framework [21].

- Relatively recently, Schmidt and Lipson [43] proposed the use of co-evolution technique to address issues such as accuracy of fitness predictor and level of approximation.

Refer to [31] for details on single and multiple surrogate management techniques.

## VII. Quality Assurance

Quality assurance is impacted by among other factors, sample selection, approximator selection, and selection of surrogate evaluation metrics. In this section we briefly cover mainly sample and evaluation metric selection issues.

If an approximate model is used for evolutionary computation, both offline and online training will be involved if the evolution is to be controlled. In this context, offline learning denotes the training process before the model is used in evolutionary computation. On the other hand, on-line learning denotes adjusting or rebuilding the model during the evolutionary process. Usually, the samples for offline learning can be generated using Monte-Carlo method; however, it has been shown in different research areas that active selection [45, 38] of the samples will improve the model quality significantly. During on-line learning, data selection is strongly related to the search process [35].

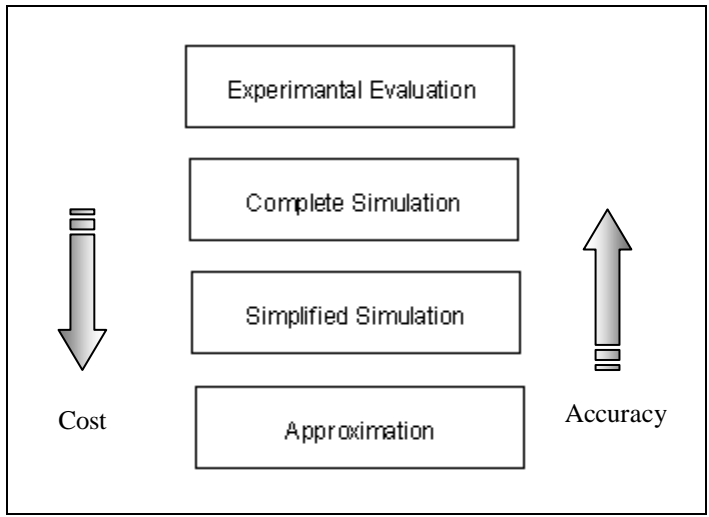

Fig. 3. Levels of fitness evaluations and their respective tradeoffs between computational expense and accuracy.

In Section IV we have briefly mentioned about a number of relatively popular approximators and have identified some of their comparative functional aspects. However, it may be noted that selection of approximator is also problem dependent among other factors.

Estimating the model quality by calculating the average approximation error after re-evaluation has been used in some research [62]. [25] has suggested a mechanism for adapting the number of individuals to be evaluated using surrogates. However, there is no clear indication as to which surrogate evaluation metric may be advantageous.

Approximation accuracy is naturally a desirable criterion for effective use of surrogates. One of the main difficulties in achieving approximation accuracy is the high dimensionality of the design space in case of most real world problems. [63] and [65] have used dimension reduction techniques to build the surrogate in a lower dimensional space to overcome this problem.

## VIII. CONCLUSIONS

Fitness approximation in evolutionary computation is a research area with major potential; but, it has not yet attracted sufficient attention in the evolutionary computation community. In the preceding sections we have presented various issues and aspects of use of approximation in EA. However, several issues still remain to be addressed for approximation based EA to be successful. Below are some of such issues:

- Theoretical research as to how EA can benefit from use of surrogates is lacking. Without a theoretical background it is hard to satisfactorily answer many of the issues raised in this paper.

- Surrogates have been used in local as well global search mechanisms in various researches. However, adequate comparative study is not available to ascertain which one is more beneficial.

- A number of different evolution control or model management techniques are available in the literature, However, still no concrete logic exists that can guide the choice of a particular model management technique over another.

- Research is lacking in the area of surrogate assisted evolutionary algorithm (or other metaheuristics) for combinatorial optimization problems that are computationally intensive.

- Further research is required in the area of surrogate assisted EA for problem domains involving variable input dimensions and dynamic optimization.

Majority of the researches available in the literature uses benchmark test functions to establish the efficacy of the proposed methods involving surrogate assisted evolutionary algorithms. However, it is important to test these methods on real world expensive optimization problems to realize their true potential or lack thereof.